\documentclass[runningheads]{llncs}

\usepackage{graphicx}
\usepackage{url}
\usepackage{graphicx}
\usepackage{amsmath}
\usepackage{multirow}
\usepackage{amssymb}

\usepackage{color}
\definecolor{yellow}{rgb}{0.863,0.741,0.137}
\definecolor{black}{rgb}{0,0,0}

\newcommand{\sgn}{\mathop{\mathrm{sgn}}}
\newcommand{\terntan}{\mathop{\mathrm{ternTanh}}}
\newcommand{\tern}{\mathop{\mathrm{tern}}}

\begin{document}

\mainmatter  

\title{TernaryNet: Faster Deep Model Inference without GPUs for Medical 3D Segmentation using Sparse and Binary Convolutions}
\titlerunning{TernaryNet: Faster Deep Model Inference w/o GPUs for 3D Segmentation}
\author{
Mattias P. Heinrich\inst{1}
\and Max Blendowski\inst{1}
\and Ozan Oktay\inst{2}}

\institute{
Institute of Medical Informatics, University of L\"{u}beck, Germany\\
\and Biomedical Image Analysis Group, Imperial College London, UK\\
\email{heinrich@imi.uni-luebeck.de}, \url{www.mpheinrich.de}}

\authorrunning{M.P. Heinrich, M. Blendowski and O. Oktay}   
\maketitle

\begin{abstract}

Deep convolutional neural networks (DCNN) are currently ubiquitous in medical imaging. While their versatility and high quality results for common image analysis tasks including segmentation, localisation and prediction is astonishing, the large representational power comes at the cost of highly demanding computational effort. This limits their practical applications for image guided interventions and diagnostic (point-of-care) support using mobile devices without graphics processing units (GPU).
We propose a new scheme that approximates both trainable weights and neural activations in deep networks by ternary values and tackles the open question of backpropagation when dealing with non-differentiable functions. Our solution enables the removal of the expensive floating-point matrix multiplications throughout any convolutional neural network and replaces them by energy and time preserving binary operators and population counts. 
Our approach, which is demonstrated using a fully-convolutional network (FCN) for CT pancreas segmentation leads to more than 10-fold reduced memory requirements and we provide a concept for sub-second inference without GPUs. Our ternary approximation obtains high accuracies (without any post-processing) with a Dice overlap of 71.0\% that are statistically equivalent to using networks with high-precision weights and activations. We further demonstrate the significant improvements reached in comparison to binary quantisation and without our proposed ternary hyperbolic tangent continuation.
We present a key enabling technique for highly efficient DCNN inference without GPUs that will help to bring the advances of deep learning to practical clinical applications. It has also great promise for improving accuracies in large-scale medical data retrieval.
 
\end{abstract}

\section{Introduction}
\label{intro}
Deep convolutional neural networks (CNNs) have been shown to substantially improve common image analysis tasks in computer vision and (bio-)medical imaging. They have in particular advanced research in automatic segmentation and image classification. Dense prediction based on fully-convolutional network (FCN) architectures \cite{long2015fully} enables very accurate voxel-wise segmentation by a single forward pass of the input image through a trained CNN architecture \cite{gibson2017towards}. However, FCNs also come with tremendous demand for memory and computational resources that can rarely be satisfied in clinical scenarios in particular when envisioning a mobile application of computer-assisted diagnosis and interventions. Furthermore, the translation of deep learning into interactive clinical workflows will require processing times of few seconds, which up-to date were only achievable using power-demanding GPUs. Surprisingly little research has been undertaken in deep learning for medical image analysis that attempts to limit model complexity. In this work, we address these challenges and present a new technique to advance state-of-the-art CNN and FCN approaches by introducing the \textbf{TernaryNet} -- a versatile end-to-end trainable deep learning architecture that drastically reduces computational and memory demand for inference. We achieve this goal by replacing floating point matrix multiplications with ternary convolutions (based on sparse binary kernels), with both activations and weights restricted to values of $\{-1,0,+1\}$. They can be calculated using a masked Hamming distance, a XOR / XNOR operation followed by a \texttt{popcount}, and reduce computational demand by up to a factor of 16. Our approach is not merely motivated by gains in computational performance, but also to explore the theoretical advantages of explicit sparsity promotion to reduce the risk of overfitting (as detailed in the following subsection)  and learn more plausible neural network models. Our work extends recent approaches from computer vision that relied on binary convolutions \cite{rastegari2016xnor}, ternary weight networks \cite{li2016ternary}, hashing by continuation \cite{cao2017hashnet} and our initial work on sparse binary convolutions \cite{heinrich2017briefnet}. The presented approach is to the best of our knowledge the first to use binary convolutions for semantic segmentation and the very first to propose ternary convolutions (and not only ternary weights since activations are also restricted) based on masked Hamming distances. 

The \textbf{TernaryNet} can be employed for any given image analysis task, e.g. landmark regression or image-level classification, but we chose to demonstrate its applicability to medical imaging for the automatic voxel-accurate segmentation of the pancreas in CT scans, which is a particularly demanding task. Pancreas segmentation is very important for computer assisted diagnosis of inflammation (pancreatitis) or cancer and furthermore to provide image-based navigational guidance for interventions, including endoscopy \cite{gibson2017towards}. In the following, we will motivate the use of sparse binary kernels in deep convolutional networks and discuss related work for the use of quantisation in image analysis in particular in deep networks. Section \ref{secMethod} contains the detailed explanation of ternary quantisation and convolutions. Starting with a short discussion of current work on CT pancreas segmentation, we describe our experimental setup in Section \ref{secExperiments} and compare different strategies and choices for model complexity reduction. We discuss our results, potentials for further research and future implications of our novel ternary convolution concept in Section \ref{secResults} and end with some concluding remarks.

\begin{figure}[t]
\begin{center}
   \includegraphics[width=0.44\linewidth]{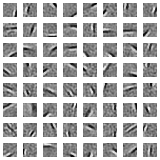}
   \includegraphics[width=0.44\linewidth]{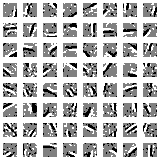}
\end{center}
   \caption{Left: Visual example of learned synthetic receptive fields (reproducing the results of \cite{lee2007efficient}) using sparse coding techniques. Right: Ternarisation of weights demonstrates the low approximation error for these naturally inspired sparse filters.}
\label{figLeeFields}
\end{figure}

\textit{Motivation for sparse binary kernels:} Convolutional neuronal networks excel in image recognition tasks by mimicking the visual cortex of mammals. The visual information is detected by photoreceptor cells and transmitted and processed using multiple layers of neurons interconnected by synapses. Computational models have the capacity to replicate these mechanisms and can furthermore represent neural activations up to extremely high numerical precision (up to 8 decimal points). However, in nature the simple structure of neural cells and environmental influences severely limit the accuracy of subtle changes in activation and in addition the need to conserve energy may lead to a sparse as possible use of neural activity. Ohlshausen \& Field \cite{olshausen1996emergence} and Lee et al. \cite{lee2007efficient} therefore established the idea of sparse coding for pattern recognition and neural networks. Those works demonstrate that powerful convolutional filters can be learned using few non-zero values by means of sparsity inducing L1 norms and a feature sign searching algorithm. Furthermore, we observe that the non-zero elements of these synthetic models of V1 cells tend to be close to values +1 and -1. Therefore, a ternary approximation of weights leads to only minor degradation of representational power (see Fig. \ref{figLeeFields}).

\textit{Related work: }Due to their computational efficiency binary codes and their comparison using the Hamming distance (which counts the number of dissimilar bits in a long binary vector) are becoming increasingly popular for demanding image analysis tasks. They have been employed for hashing based large-scale image retrieval \cite{conjeti2017hashing,zhang2015towards}, nearest-neighbour based segmentation \cite{heinrich2016multi} and image registration \cite{heinrich2013edge}. In computer vision binary descriptors are frequently used for realtime applications, e.g. tracking using BRIEF features \cite{calonder2012brief}. There are, however, also cases where binarisation led to inadequate loss in representation quality as e.g. reported for lung nodule classification in \cite{farag2010toward}. 

In our recent prior work \cite{heinrich2017briefnet}, we proposed the use of sparse binary kernels with very large receptive fields inspired by BRIEF features and dilated convolutions \cite{wolterink2016dilated,yu2015multi} that enabled highly accurate segmentations without complex network architectures. Similarly and concurrently \cite{juefei2017local} proposed local binary convolutions that are derived from local binary patterns. A key limitation of these works is, however, that their design does not allow us to automatically train non-zero elements within binary kernels. Instead, they have to be chosen once at random (with a similar manual design as proposed in \cite{calonder2012brief}). We also did not realise binary or ternary activations thus the use of efficient computations without floating point arithmetic was not possible. An alternative solution that has recently been proposed is the use of trained ternary filter weights \cite{li2016ternary,zhu2016trained}. In particular ternary weight networks \cite{li2016ternary} use a very simple, yet powerful, approximation and learning strategy based on the mild assumption of Gaussian statistics. They generalise the earlier ideas of \cite{courbariaux2016binarized,rastegari2016xnor} for learning binary weights and clearly demonstrate that ternarisation drastically reduces the accuracy gap to high precision weights. Another related approach by Liu et al. \cite{liu2015sparse} employs decomposition methods for sparsification of convolution filters and proposes a new implementation for fast sparse matrix multiplication. 

While weight quantisation has quickly matured, another important aspect that has so far been only insufficiently addressed is the quantisation or sparsification of activations. Setting approximately half of the activations to zero using a rectifying linear unit (ReLU) is common practice in deep learning. Yet more drastic quantisation e.g. using the sign function
\begin{equation}
\sgn(x):=(x\ge0\rightarrow1)\wedge(x<0\rightarrow-1)
\end{equation}
as non-linear activation leads to strong artefacts during forward passes and no gradient for backpropagation. Courbariaux et al. \cite{courbariaux2016binarized} therefore proposes an adhoc solution that employs a rectangle (boxcar) function
\begin{equation}
\partial\sgn/\partial x\approx(\vert x\vert\le1\rightarrow1)\wedge(\vert x\vert>1\le0\rightarrow0)
\label{eqAdhoc}
\end{equation}
as a replacement, which was later also used in \cite{rastegari2016xnor}. The downside of this approach is the fact that since two different functions are used during forward and backward propagation the training behaviour is ill-defined and potentially unstable. Cao et al. \cite{cao2017hashnet} propose a more justifiable approach based on the continuation of the hyperbolic tangent, which approaches the sign function with increasing slope $\beta$ in its limit:
\begin{equation}
\lim_{\beta\rightarrow\infty}\tanh(\beta x)=\sgn(x)
\label{eqContinuation}
\end{equation}
They prove the convergence of this optimisation when employing a sequence of increasing values of $\beta$ during training. They limit the use of this function to the final layer within a framework for supervised hashing. In our work, we extend this concept to a ternary hyperbolic tangent as explained in detail in the following section and apply this function as nonlinearity throughout -- for every activation -- in our deep network models.

\section{Method}
\label{secMethod}
We aim to automatically segment the pancreas in regions of interest extracted from CT volumes. For this purpose use a fully-convolutional U-Net architecture \cite{ronneberger2015u} is chosen. However,  a V-Net \cite{milletari2016v} or multi-path network will most likely lead to similarly good segmentations and would also support our findings. The U-Net model can contain several million free parameters rendering it computationally demanding and prone to overfitting. Furthermore, as common for FCN architectures an efficient inference requires an unexpectedly large amount of memory due to the use of the \texttt{im2col} operations. They are necessary to perform multi-channel convolutions of all elements in the feature maps in parallel using matrix multiplications between activations of preceding layers with a current filter bank \cite{jia2014learning}. 
We propose a ternary quantisation of weights and activations that is generic and therefore applicable to reduce complexity for any (convolutional) neural network architecture including FCNs.

\textit{Ternary weights:} In order to limit the memory demand, reduce model complexity and enable inference of CNNs in practical clinical environments, it is desirable to reduce the precision of both activations and weights. Following the recent work of Li et al. \cite{li2016ternary}, we aim to find the best approximation to the filter weights $\mathbf{W}\approx\alpha\mathbf{\tilde{W}}$ where $\alpha$ describes a (floating point) scaling parameter and $\mathbf{\tilde{W}}$ consists of only ternary values $\{-1,0,1\}$. It is shown in \cite{li2016ternary} that the minimal quantisation error can be obtained by calculating:
\begin{equation}
\mathbf{\tilde{W}}_i=\begin{cases}+1&\text{ if }W_i>\Delta\\0&\text{ if }\vert W_i\vert\le\Delta\\-1&\text{else}\end{cases}\text{ with }\Delta=\frac{0.7}{n}\sum_{i=1}^n\vert W_i\vert
\label{eqTernaryWeight}
\end{equation}
and $\alpha=\frac{1}{n_{\Delta}}\sum_i\vert\tilde{W}_i\vert\vert W_i\vert$ with $n_{\Delta}=\sum_i\vert\tilde{W}_i\vert$. When employing quantised weights during the training of a network using stochastic gradient descent with mini-batches (i.e. in virtually any case of deep learning) it is strongly advisable \cite{courbariaux2016binarized} to accumulate gradient updates with full-precision (while using $\mathbf{\tilde{W}}$ for both forward and backward passes), otherwise they would usually not exceed the threshold (according to Eq. \ref{eqTernaryWeight}) necessary to flip individual bits. This simple and straightforward ternary weight approximation already yields excellent accuracies for classification tasks (only  3.6\% lower top-1 scores for ImageNet compared to full-precision networks \cite{li2016ternary}).

\begin{figure}
\begin{center}
   \includegraphics[height=4cm]{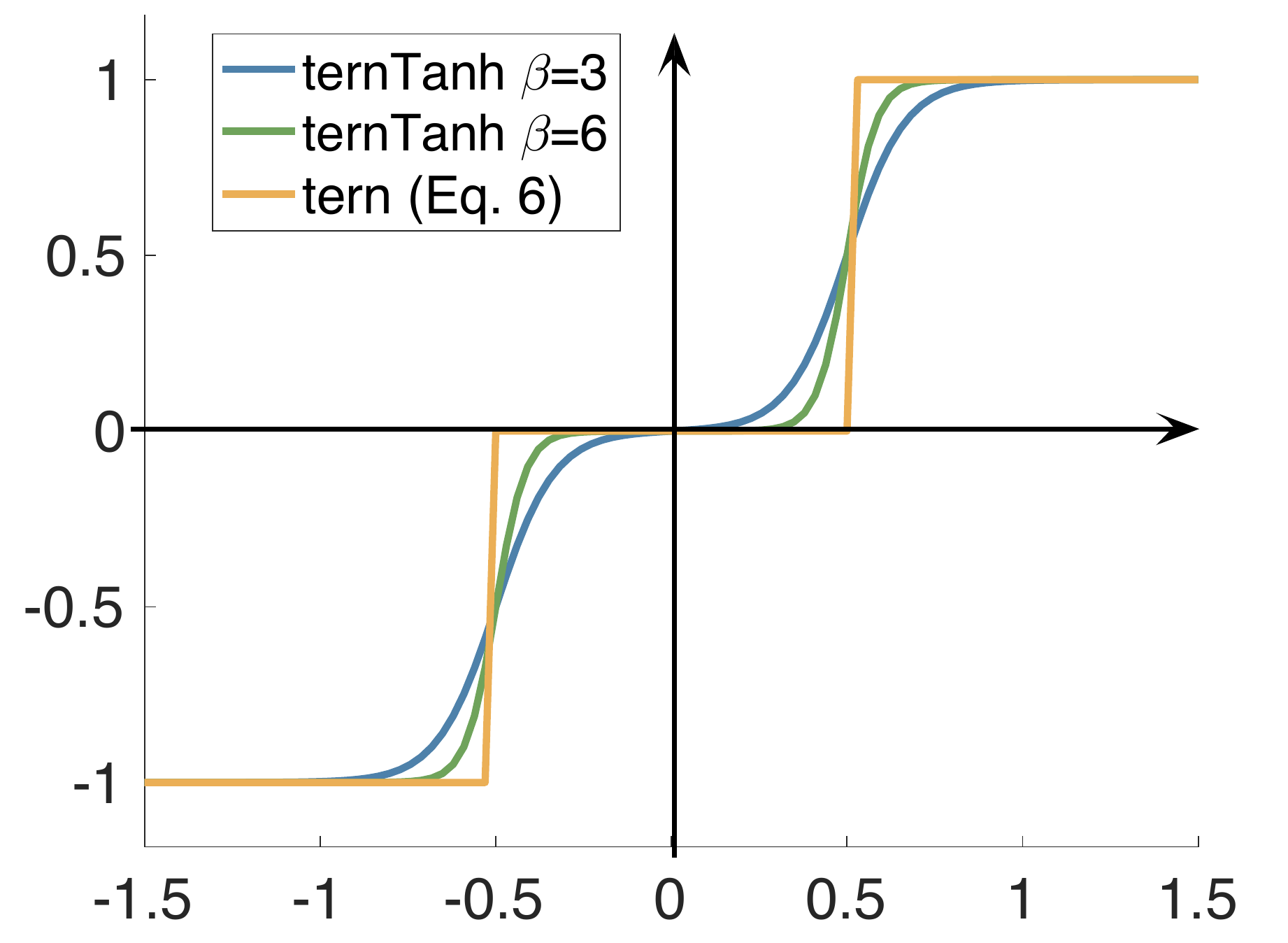}
   \includegraphics[height=4cm]{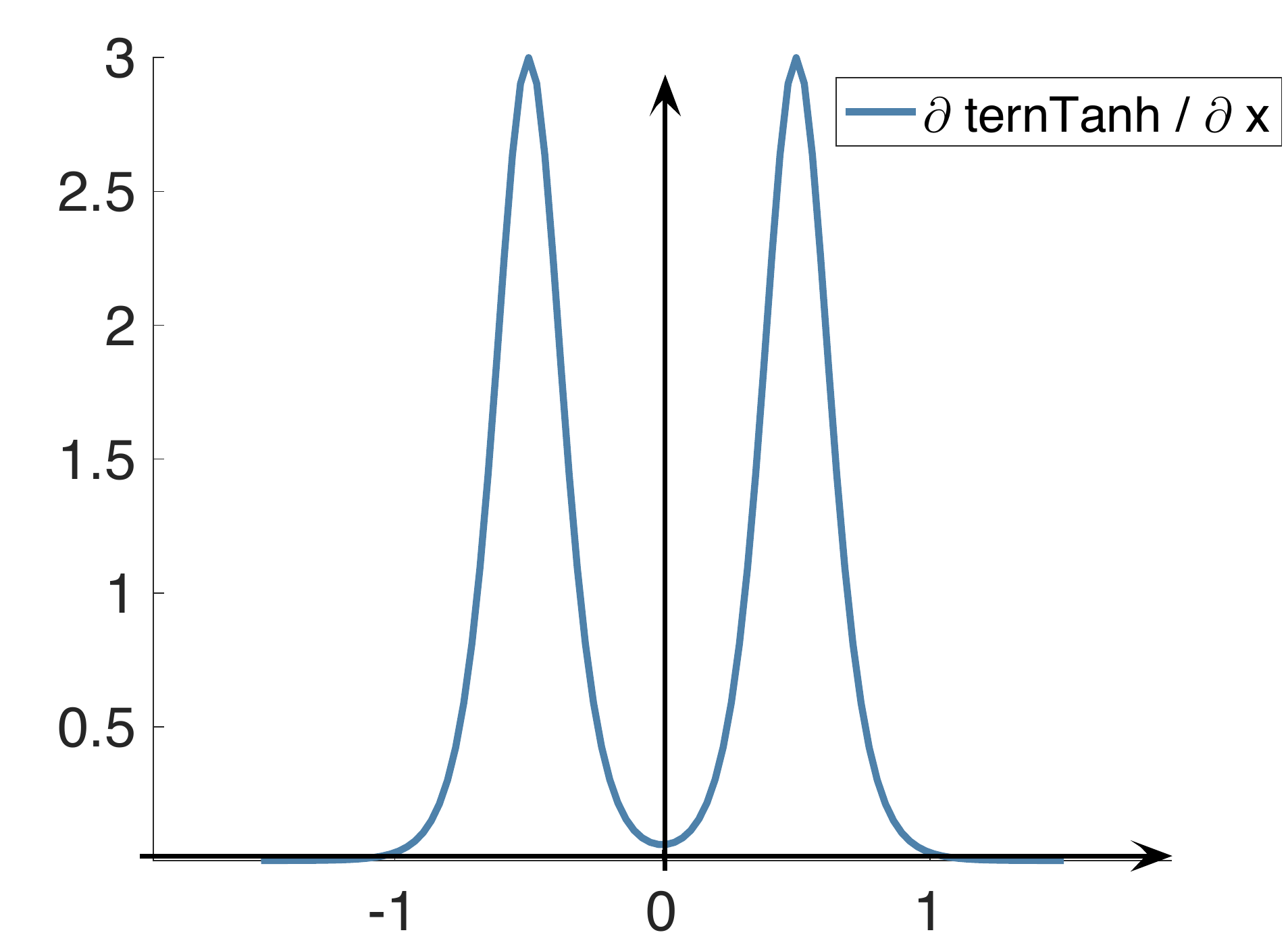}
\end{center}
   \caption{Visualisation of proposed ternary hyperbolic tangent as defined in Eq.~\ref{eqTHT} showing varying $\beta$ values for increasing steepness of slopes. The analytical derivative of our new nonlinearity is shown for $\beta=3$ on the right.}
\label{figTHT}
\end{figure}

\textit{Ternary activations:} The use of ternary weight approximations alone, however, cannot reduce the huge memory and computational demand required to store and process intermediate feature maps, since the resulting activations will still be full precision. The key contribution of our work is therefore the introduction of a new activation function that enables an accurate ternarisation of intermediate features in a neural network, which we coin \textbf{ternary hyperbolic tangent}. This proposed function $\terntan(x)$ combines two hyperbolic tangents to form plateaus around zero and beyond +1 and -1:
\begin{equation}
\terntan(x) = \frac{1}{2}\tanh(2\beta x-\beta) - \frac{1}{2}\tanh(-2\beta x-\beta)
\label{eqTHT}
\end{equation}
In contrast to a sign function the ternary hyberbolic tangent is fully differentiable and can therefore be used without custom changes to the learning procedure of deep networks. The parameter $\beta$ controls the slope and can be varied throughout the process of learning. In earlier iterations it is beneficial to use smaller values for $\beta$ to enable sufficient gradient flow and avoid ``dying'' neurons. Eventually, we aim for a discrete step function $\tern(x)$ that can be defined as:
\begin{equation}
\tern(x)=\begin{cases}+1&\text{ if }x>0.5\\0&\text{ if }\vert x\vert\le0.5\\-1&\text{else}\end{cases}
\label{eqTernaryAct}
\end{equation}
Similar as above for the binary case covered in \cite{cao2017hashnet} the following continuation holds true (see Fig. \ref{figTHT} for visual example):
 \begin{equation}
 \lim_{\beta\rightarrow\infty}\terntan(\beta x)=\tern(x)
 \label{eqTHTlim}
 \end{equation}

\textit{Ternary convolutions and complexity analysis:} In combining both ternary weights and ternary activations, we can realise important avoidance of time-consuming floating point multiplications, which were at the core of classical deep learning architectures. In \cite{courbariaux2016binarized,rastegari2016xnor} the idea of replacing full precision inner products of an input tensor $\mathbf{I}$ and a filter bank $\mathbf{W}$ by boolean operations and bit counting (population count) was explored for binary valued operands, i.e. $\mathbf{I},\mathbf{W}\in\{-1,+1\}^c$, where $c$ denotes the size of a kernel (including both spatial extend and number of features). It is straightforward to show that a matrix multiplication and its inner-products can be efficiently calculated in the Hamming space:
\begin{equation}
\mathbf{I}_i\mathbf{W}_j = c-2\Xi\{\mathbf{I}_i\oplus\mathbf{W}_j\}
\label{eqBinaryConv}
\end{equation}
where $\oplus$ defines an exclusive OR (XOR) operator and $\Xi$ a bit-count over the $c$ bits in the rows of $\mathbf{I}$ and $\mathbf{W}$. Modern CPUs, FPGAs or embedded SoCs all contain instructions for efficiently calculating population counts of 64-bit strings in few cycles (using AVX extensions Intel CPUs achieve a throughput of 0.5 cycles \cite{mula2018faster}). This means that each bit-count replaces 64 floating point multiplications and additions. Even when considering the highly optimised fused multiply addition (FMA) instructions on 256 bit wide registers (\texttt{mm256-fmadd-ps}), which are employed on modern Intel CPUs and that can process 8 packed FMAs in parallel in 0.5 cycles, we can gain a speed up of a factor of 8. When considering equal power consumption (floating point operations require more complex logic) the improvements are even much higher.

\begin{figure}
\begin{center}
   \includegraphics[width=0.9\linewidth]{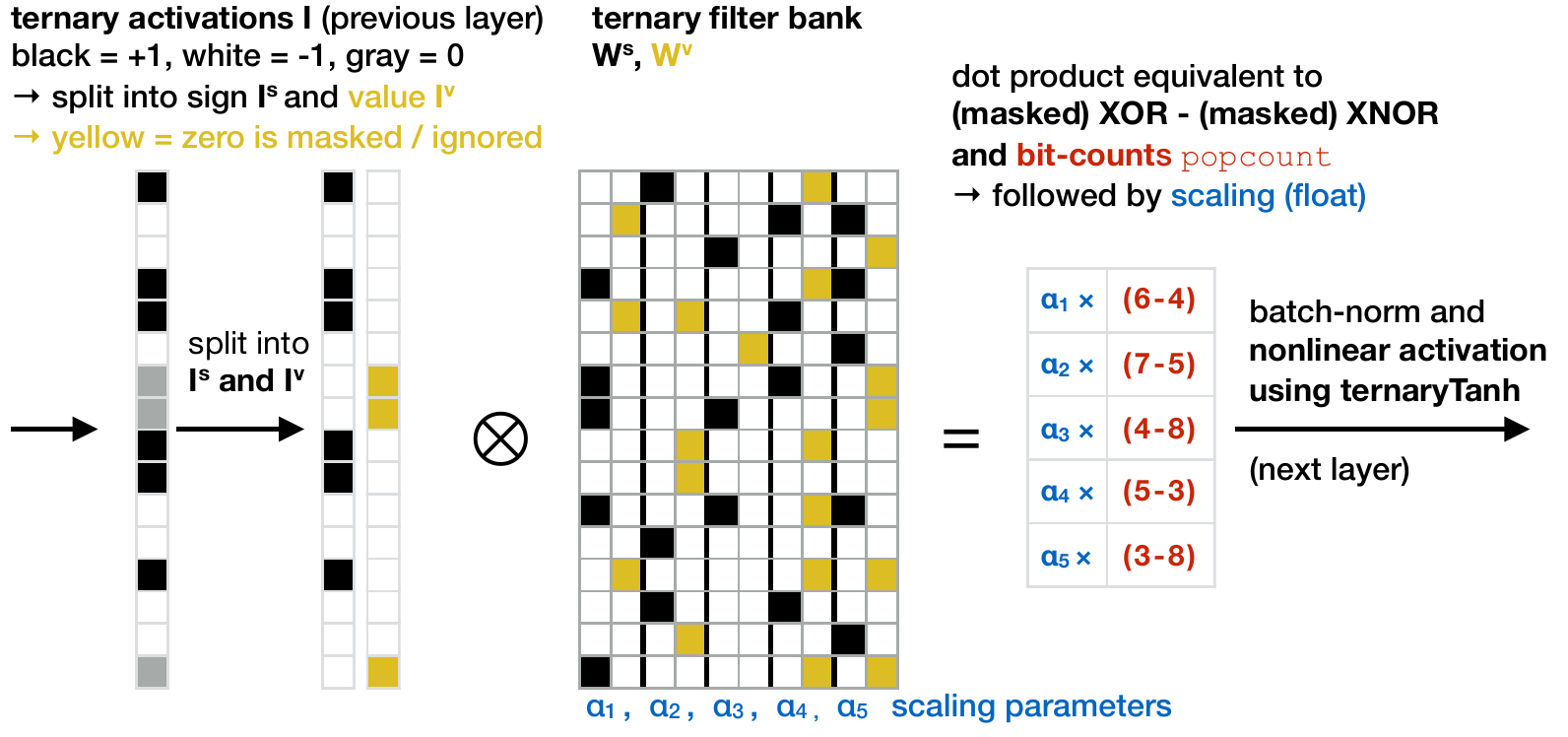}
\end{center}
   \caption{Visual example for the computation of ternary convolutions without floating point operations. Ternary values are encoded by sign and value, i.e. +1 $\rightarrow$ (\textcolor{black}{$\blacksquare$},$\Box$), -1 $\rightarrow$ ($\Box$,$\Box$) and 0 $\rightarrow$ ($\Box$,\textcolor{yellow}{$\blacksquare$}). The approximation for a ternary filter bank provides scaling parameters $\alpha$ see below Eq.~\ref{eqTernaryWeight}. Ternary convolutions can be computed by masked XOR and XNOR operators followed by a bit-count according to Eq.~\ref{eqTernaryConv}. The output is batch normalised and passed on to the nonlinearity visualised in Fig.~\ref{figTHT}. }
\label{figConcept}
\end{figure}

Since previous work on binary quantisation of deep learning architectures \cite{courbariaux2016binarized,rastegari2016xnor} has led to severely reduced accuracy of 12-20\% for image classification tasks, we aim to extend the concept of bit-counting as replacement for matrix multiplications to ternary valued networks with $\mathbf{I},\mathbf{W}\in\{-1,0,+1\}^c$. As shown in Fig.~\ref{figConcept}, we can store ternary tensors using 2 bits per entry that encode the sign and value respectively. We denote these two tensors as $\mathbf{I}^s, \mathbf{I}^v\in\{0,1\}^c$ and $\mathbf{W}^s, \mathbf{W}^v\in\{0,1\}^c$.  The inner-product calculation can then be realised using two bit-counts in Hamming space:
\begin{equation}
\mathbf{I}_i\mathbf{W}_j = \Xi\{\overline{(\mathbf{I}_i^s\oplus\mathbf{W}_j^s)}\&(\mathbf{I}_i^v+\mathbf{W}_j^v)\} - \Xi\{(\mathbf{I}_i^s\oplus\mathbf{W}_j^s)\&(\mathbf{I}_i^v+\mathbf{W}_j^v)\}
\label{eqTernaryConv}
\end{equation}
Here, \& defines an AND operator, $+$ the boolean OR and $\overline{A\oplus B}$ the negated XOR. A more intuitive interpretation is that all operations involving a zero value are excluded and the first part of the equation calculates all positives elements of a dot product, i.e. $+1\cdot+1$ and $-1\cdot-1$, while the second part subtracts the number of times an opposing sign multiplication occurs. The complete concept of an individual building block for ternary convolutions in deep networks is show in Fig.~\ref{figConcept}. In practice further speed-ups (halving the number of bit-counts) are possible when training the weight quantisation to follow a specified degree of sparsity, e.g. by replacing the rule derived in Eq. \ref{eqTernaryWeight} and specify $\Delta$ so that in each kernel exactly 50\% of entries are zero.

In summary, each module in our proposed \textbf{TernaryNet} architecture comprises a ternary approximation of filter weights together with a ternarisation of activations to enable low-power, high-speed ternary convolutions without floating point operations. During training both weight updates for mini-batch optimisation and the activations using the new ternary hyperbolic tangent $\tern$ are kept at full precision to enable gradient flow and precise learning. By extending the strategy of \cite{cao2017hashnet} to ternary activations and applying a continuously increasing slope $\beta$ during training, the network learns to cope with sparse and quantised activations, which is vital in order to avoid diverging objectives between training and testing. Batch-normalisation layers \cite{ioffe2015batch} are inserted between ternary convolutions and activations to accelerate the learning process and keep a zero mean of feature responses as well as an approximately unit normal distribution to ensure the nonlinearity is not easily saturated. A trained model can be stored using only 2 bits per weight and one (full-precision) scalar weighting value per feature channel -- reducing the required memory by more than an order of magnitude. During model inference on unseen data, we employ a hard quantisation of Eq.~\ref{eqTHT} and thereby enable the use of Hamming distances for ternary convolutions. It is important to note that all common architectural design choices of modern deep networks, such as skip connections \cite{ronneberger2015u}, dilated kernels \cite{heinrich2017briefnet,yu2015multi} or dense feature concatenation \cite{gibson2017towards,huang2016densely} are useable with ternary convolutions. 

\section{Experiments}
\label{secExperiments}
To demonstrate the usefulness of \textbf{TernaryNets} for highly efficient medical image analysis, we explore the dense prediction (semantic segmentation) of the pancreas in CT. The extension of our model to multi-organ labelling is straightforward. Providing image-guidance for interventional tasks relies on fast inference executed on common clinical workstations or even mobile devices. We therefore also analyse in detail the computational operations and memory requirements in our experiments. The highly variable shape and a relatively poor contrast of the pancreas as well as confusable neighbouring abdominal anatomies make this segmentation very difficult. Therefore, networks with large receptive fields are required to robustly capture sufficient regional context, while at the same time an automatic method should delineate local objects boundaries accurately and avoid over-segmentation of similar neighbouring structures within the field-of-view. Our experiments are based on the public NIH dataset that was described in \cite{roth2015deeporgan}. It comprises 82 high-resolution CT scans along with accurate manual segmentations for training and validation. 

\textit{Comparison to State-of-the-Art: } Several approaches have been evaluated in the last few years on the NIH dataset and a similar corpus of abdominal CT scans (the BCV challenge data described in \cite{xu2016evaluation}). Accuracies for pancreas segmentation without CNNs are often relatively low, e.g. overlap scores of 40\% and 49\% have been reported for two different multi-atlas techniques in \cite{xu2015efficient}. Employing discrete registration within multi-atlas label fusion \cite{heinrich2015multi} improved accuracies for pancreas segmentation to 74\% Dice, ranking first at the MICCAI 2015 BCV challenge. The approach of \cite{larsson2017robust} reached 60\% overlap within the same challenge by combing registration based localisation with deep CNNs. Roth et al. achieved a Dice score of 71\% \cite{roth2015deeporgan} on the NIH dataset when combining supervoxel based deep region regression with CNN patch classification and could further improve their accuracy to 78\% \cite{roth2016spatial} using holistically nested networks together with random forest classifiers. Very recently, Zhou et al. \cite{zhou2017fixed} achieved an astonishing performance of 82\% on the NIH data by training an iterative sequence of multiple (coarse-to-fine) deep CNNs. The use of densely connected layers within a V-Net architecture (Dense V-Net \cite{gibson2017towards}) resulted in a Dice overlap of 66\% (on both NIH and BCV datasets), which is also the only of the mentioned deep learning approaches that did not rely on a combination of classifiers or registration. In our own previous work \cite{heinrich2017briefnet} we reached 65\% Dice for the BCV dataset using (untrained) sparse binary convolutions that enabled huge receptive fields but no binary (or ternary) convolutions.

\textit{Baseline model}: Our aim is not necessarily to surpass current state-of-the-art accuracies, but to demonstrate and analyise the effects of network model quantisation. We therefore employ a four-level fully-convolutional U-Net architecture \cite{ronneberger2015u} as an exemplary baseline. To fairly assess the influence of binarisation and ternarisation, we employ the same number of channels and convolution filters for all compared models and hyperbolic tangents (except for the final prediction layer) as baseline activation function. Table \ref{tabUnet} lists the details of the chosen architecture, including the number of floating point operations (FMAs) required per layer. The resulting receptive field of this network is 36 voxels. Using floating-point precision, the network requires  2.6 million weights and thus 10.6 MBytes of storage for the model weights. During training the model requires more than 5 GBytes of memory (using a mini-batch size of 10), for inference this can be reduced to approximately 1 GByte.  

\begin{table}
\caption{Description of baseline U-Net model. Number of million fused multiply add (floating point operations) is given as MFlops. To reduce the number of trainable parameters the convolutions in the lowest resolution level are 1$\times$1. Outgoing and incoming skip connections are noted in the last column.}
\label{tabUnet}     
\begin{tabular}{l|ccc|cc}
\hline\noalign{\smallskip}
Layer & (Out)-Size & Kernel &\# Channels & MFlops & Skip \\
\noalign{\smallskip}\hline\noalign{\smallskip}
 Input & 236$\times$172$\times$15 &&&&\\
\#1 Conv3D & 234$\times$170&3$\times$3$\times$15 &32&172&\\
\#2 Conv2D & 232$\times$168&3$\times$3 &64&718&$\rightarrow$\#13\\
\#3 Conv2D & 228$\times$164&3$\times$3 &64&345&\\
 AvgPool2D & 114$\times$82&2$\times$2 &&&\\
\#4 Conv2D & 112$\times$80&3$\times$3 &128&661&$\rightarrow$\#11\\
\#5 Conv2D & 108$\times$76&3$\times$3 &128&303&\\
 AvgPool2D & 54$\times$38&2$\times$2 &&&\\
\#6 Conv2D & 52$\times$36&3$\times$3 &256&552&$\rightarrow$\#9\\
\#7 Conv2D & 52$\times$36&1$\times$1 &256&31&\\
 AvgPool2D & 26$\times$18&2$\times$2 &&&\\
\#8 Conv2D & 26$\times$18&1$\times$1 &256&31&\\
 Upsample2D & 52$\times$38&2$\times$2 &&&\\
\#9 Conv2D & 50$\times$34&3$\times$3 &256&2005&$\#6\rightarrow$\\
\#10 Conv2D & 48$\times$32&3$\times$3 &128&453&\\
 Upsample2D & 96$\times$64&2$\times$2 &&&\\
\#11 Conv2D & 94$\times$62&3$\times$3 &128&1719&$\#4\rightarrow$\\
\#12 Conv2D & 92$\times$60&3$\times$3 &64&407&\\
 Upsample2D & 184$\times$118&2$\times$2 &&&\\
\#13 Conv2D & 180$\times$116&3$\times$3 &64&1583&$\#2\rightarrow$\\
\#14 Conv2D & 176$\times$110&3$\times$3 &64&770&\\
 Prediction & 176$\times$110&3$\times$3 &2&2&\\
\noalign{\smallskip}\hline
\end{tabular}
\end{table}

\textit{Compared models: }We have analysed in total seven variants of our baseline network to explore the effect of sparsity and quantisation to both activations and filter weights. Starting from the same baseline model, we define our \textbf{TernaryNet} by approximating weights using the ternary quantisation of Eq.~\ref{eqTernaryWeight} as proposed in \cite{li2016ternary}. During training we varied the value of $\beta$ in Eq.~\ref{eqTHTlim} linearly (and evenly with epochs) from 3.0 to 8.0 following the principal of continuation of \cite{cao2017hashnet}. The variant \textit{no continuation} uses a fixed $\beta=3$ for all epochs. To quantify whether our approach succesfully reduces quantisation loss, we also compare a variant \textit{without quantisation} that does not realise ternary convolutions. For binary convolutional networks (termed XNORnet \cite{rastegari2016xnor}, see Eq.~\ref{eqBinaryConv}), we explore the adhoc gradient approximation according to the seminal work in \cite{courbariaux2016binarized}. As alternative, we adopt the continuation (see Eq.~\ref{eqContinuation}) for a classical $\tanh$ nonlinearity. Finally, the full-precision network is compared with ReLU activations for completeness.

\textit{Data processing: }We resampled the original scans of the NIH dataset that had axial dimensions of 512$\times$512 and 181--466 slices with thicknesses between 0.5mm --1.0mm to isotropic voxel sizes of 1.0mm$^3$. We then performed a region-of interest cropping with bounding boxes of dimensions 194$\times$122$\times$138 around the pancreas, yielding an approximate density of 2\% for organ voxels (and 98\% background). There exist several accurate algorithms that automatically predict bounding boxes and/or organ locations, e.g.\cite{urschler2018integrating,xu2016bodywise}, which could be employed for this task so it was considered out of scope for our study. Subsequently, we applied a zero mean unit variance transformation on the cropped CT volumes. Following related work on pancreas segmentation using CNNs \cite{roth2015deeporgan,zhou2017fixed}, we employ only 2D convolutions, but provide a stack of several neighbouring slices (15 in our experiments) to each network. The output for each stack will be a probabilistic map of foreground and background probabilities for the given central slice. No form of post-processing is employed, which could potentially further increase accuracy, but also influences the assessment of differences across methods. 

\textit{Training and implementation details: }We use a mini-batch size of 10 and Adam with an initial learning rate of 0.0025. Each network is trained for 40 epochs with 150 iterations (1500 3D input stacks) each. Since, we encountered a huge class imbalance, we use a weighted cross-entropy loss with 0.5 for background and 2.5 for organ pixels. We trained 5 separated folds of training and validation splits using 65-66 scans for training and 16-17 for testing. The derivatives of our ternary activation and the equivalent binary $\tanh(x)$ can be found analytically (using automatic differentiation), for the adhoc approximation of binary activations in Eq.~\ref{eqAdhoc} we implemented a custom forward and backward pass. When approximating filter kernels, we keep a copy of the full precision weights, perform the quantisation before forward pass and restore the original values after the backward pass and before calling the optimiser that performs a gradient step. To enable a reproduction of our results and further research, our pytorch implementation and pre-trained models will be made publicly available after submission at \url{https://github.com/mattiaspaul/TernaryNet}.

\begin{figure}
\begin{center}
   \includegraphics[width=0.32\linewidth]{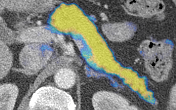}
   \includegraphics[width=0.32\linewidth]{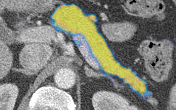}
   \includegraphics[width=0.32\linewidth]{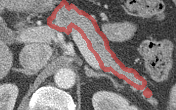}\\
   \includegraphics[width=0.4\linewidth]{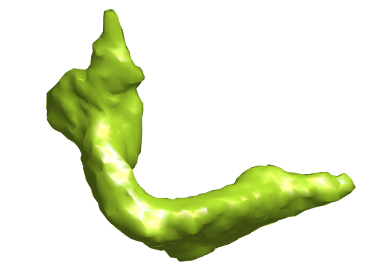}
   \includegraphics[width=0.5\linewidth]{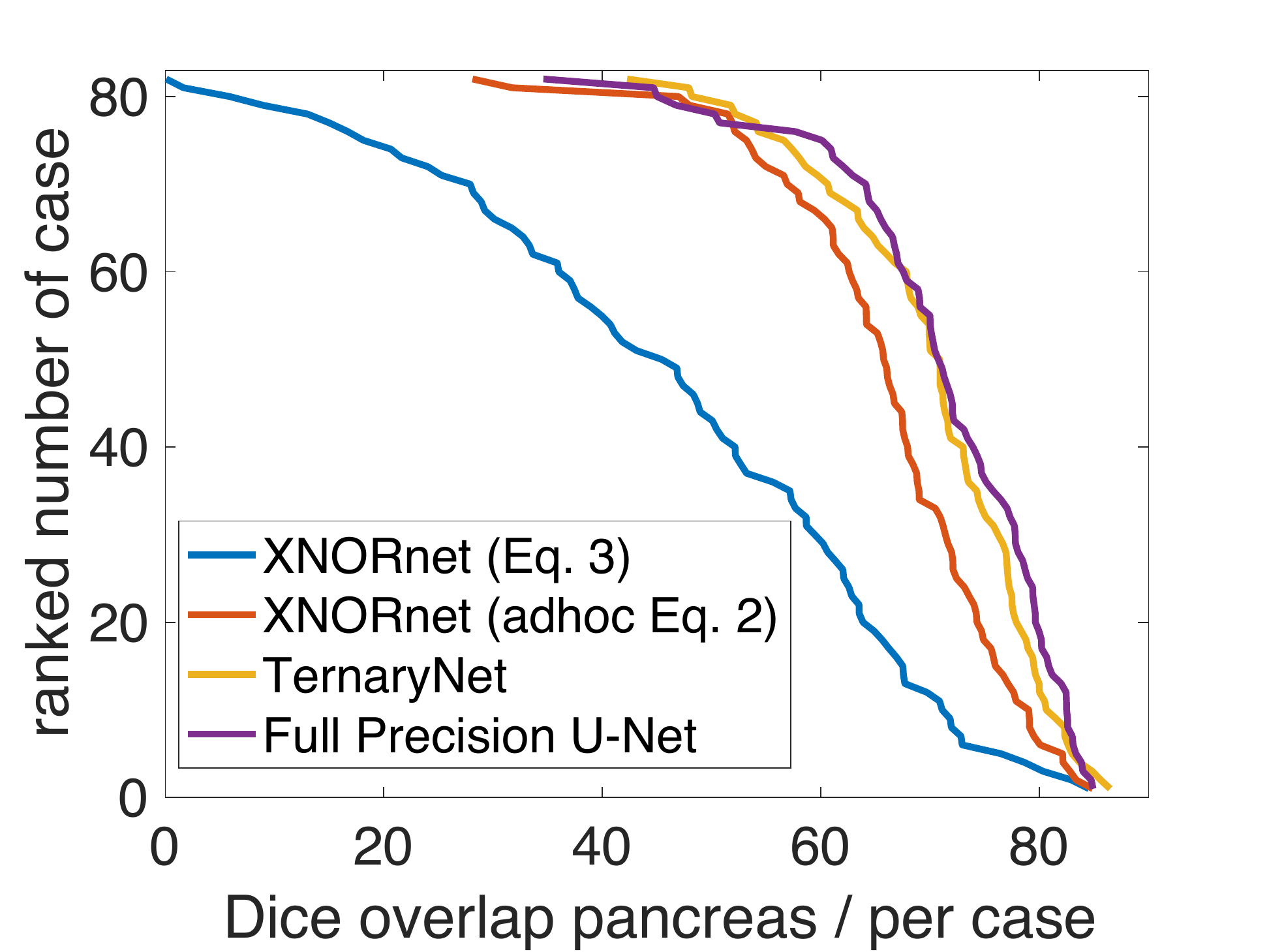}
   
\end{center}
   \caption{Top row: A visual comparison of case \# 12 of the NIH demonstrates small but significant advantages of the ternary quantisation (middle) over the better performing adhoc binary activation and quantisation, which over-segments a neighbouring structure (left). Our approach better matches the manual segmentation (right). Bottom row: 3D visualisation of our segmentation shows a very smooth surface (left). Ranked (sorted) Dice score compared across methods demonstrate that the full-precision model is not significantly better than our heavily quantised TernaryNet. Both Binary XNORnet variants perform inferior. }
\label{figVisualResults}
\end{figure}

\section{Results and Discussion}\label{secResults}
The performance of the seven compared models is evaluated quantitatively in terms of Dice overlap between automatic prediction (without further post-processing) and manual annotation. Average Dice values (and standard deviations) are compared in Table~\ref{tabResults} alongside with statistical significance tests and memory usage for model parameters. 

It can be seen that our proposed ternary convolutions perform on par with full-precision networks reaching an average Dice of 71.0\%. This demonstrate the robustness and high accuracy of our proposed ternary quantisation scheme. The results are also comparable to a number of recent deep learning approaches that all relied on full-precision and thus much larger and more complex models. When replacing the $\tanh(x)$ nonlinearity with a ReLU in the full-precision model its accuracy can be further improved by 3.8\%. However, the presumptions that symmetric activations are nowadays unsuitable to reach high accuracy has been refuted. Possibly, because the U-Net and similar architectures enable a very good backwards flow of gradients through their skip connections. The performance of binary quantisation is significantly lower than our approach. This is in particular evident for the variant that uses an analytically differentiable activation. We assert that this underlines the importance of sparse activations, which can contain a larger number of zero values -- a key feature of our new nonlinearity. Sparse intermediate feature maps enable the network to adapt certain filter banks to specific subproblems while being unaffected by pertubations of unrelated data.

\begin{table}
\caption{Dice overlap measures of pancreas for 82 CT scans (5-fold cross-validation). Paired t-tests are performed for significance analysis against \textbf{TernaryNet}, where (-) indicates that our method performed significantly better.}
\label{tabResults}  
\begin{tabular}{l | r l c | c}
\hline\noalign{\smallskip}
Architecture & \textbf{Avg. Dice} & stddev & $p$-value & weight memory  \\
\noalign{\smallskip}\hline\noalign{\smallskip}
\textbf{Binary XNORnet}& & &&\\
(continuation Eq.~\ref{eqContinuation}) &48.4\% &$\pm$20.1& $\ll$0.001 (--)& 0.33 MBytes\\
(adhoc gradients Eq.~\ref{eqAdhoc}) & 66.9\%  &  $\pm$10.5& 0.01(--)& 0.33 MBytes\\
\noalign{\smallskip}\hline\noalign{\smallskip}
\textbf{TernaryNet} &  & &&\\
(using $\beta\to\infty$ in Eq.~\ref{eqTHT})&\textbf{71.0\%} & \textbf{$\pm$9.5}& \textbf{*} &  \textbf{0.66 MBytes}\\
(without quantisation) & 71.8\% & $\pm$10.7 & 0.60 (o)&  0.66 MBytes\\
(no continuation in training) & 56.3\%  & $\pm$19.3& $\ll$0.001 (--)& 0.66 MBytes \\
\noalign{\smallskip}\hline\noalign{\smallskip}
Full Precision \textbf{U-Net} & 71.9\%  & 10.2&0.54 (o)&10.6 MBytes\\
ReLU instead of tanh & 75.7\%  & 9.0& 0.001 (+)&10.6 MBytes\\
\noalign{\smallskip}\hline
\end{tabular}
\end{table}

Training one entire model (within 40 epochs) requires about 15 minutes on an NVIDIA Titan Xp. Inference of the full precision network on a CPU takes about 80 seconds. When employing a customised OpenCL implementation for Hamming distance calculation (used for ternary convolutions in Eq.~\ref{eqTernaryConv}), we estimated inference times of 5-7 seconds using a dual-core mobile CPU. This represents a more than 10$\times$ speed-up through our contributions. Further speed-ups can be gained by reducing the number of parameters in the expanding path and skipping every other slice in a 3D volume (and interpolating in between) or adjusting the ternary weight quantisation to increase sparsity and reduce the number of population counts.

\begin{figure}
\begin{center}
   \includegraphics[width=0.48\linewidth]{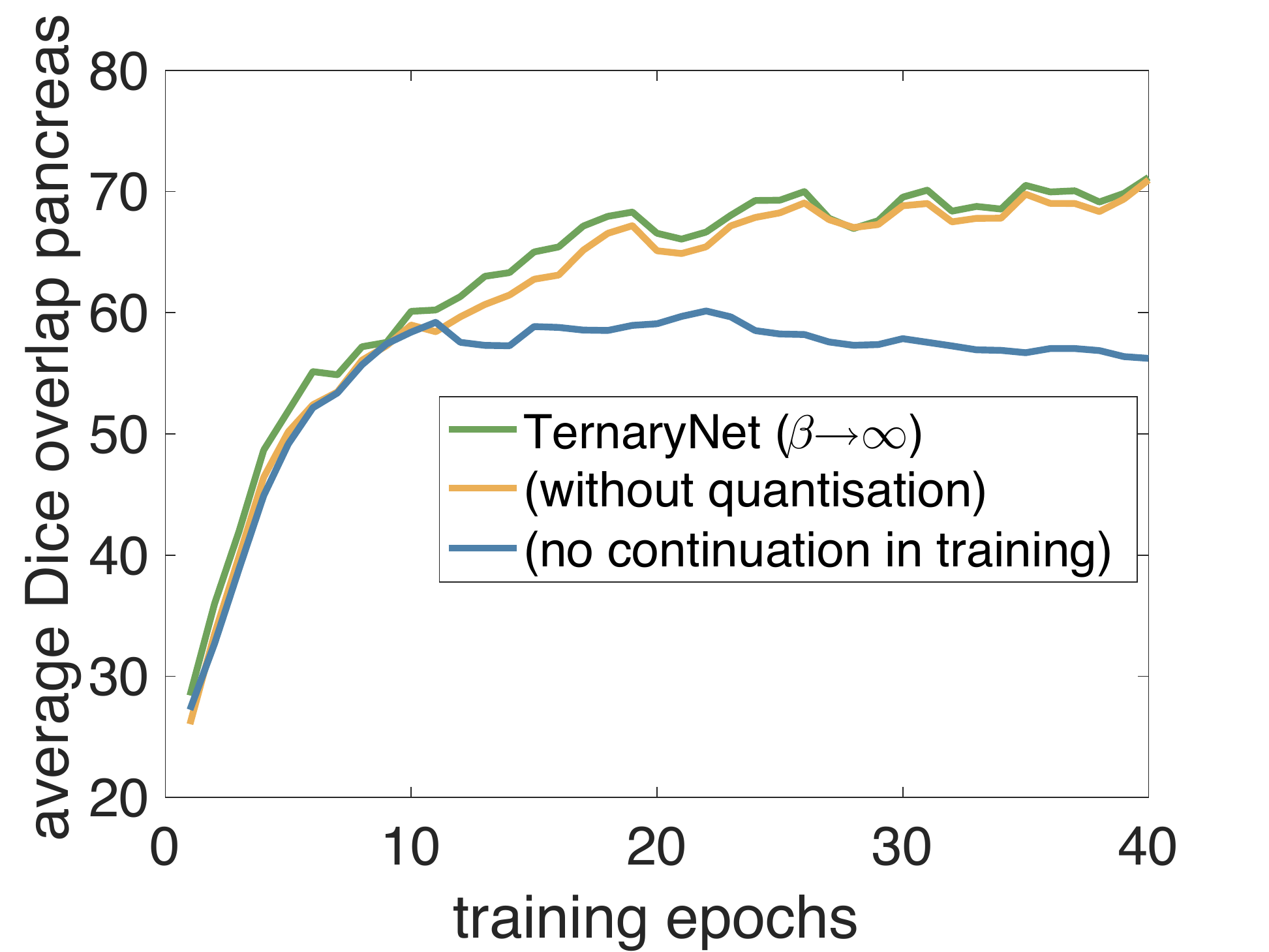}
   \includegraphics[width=0.48\linewidth]{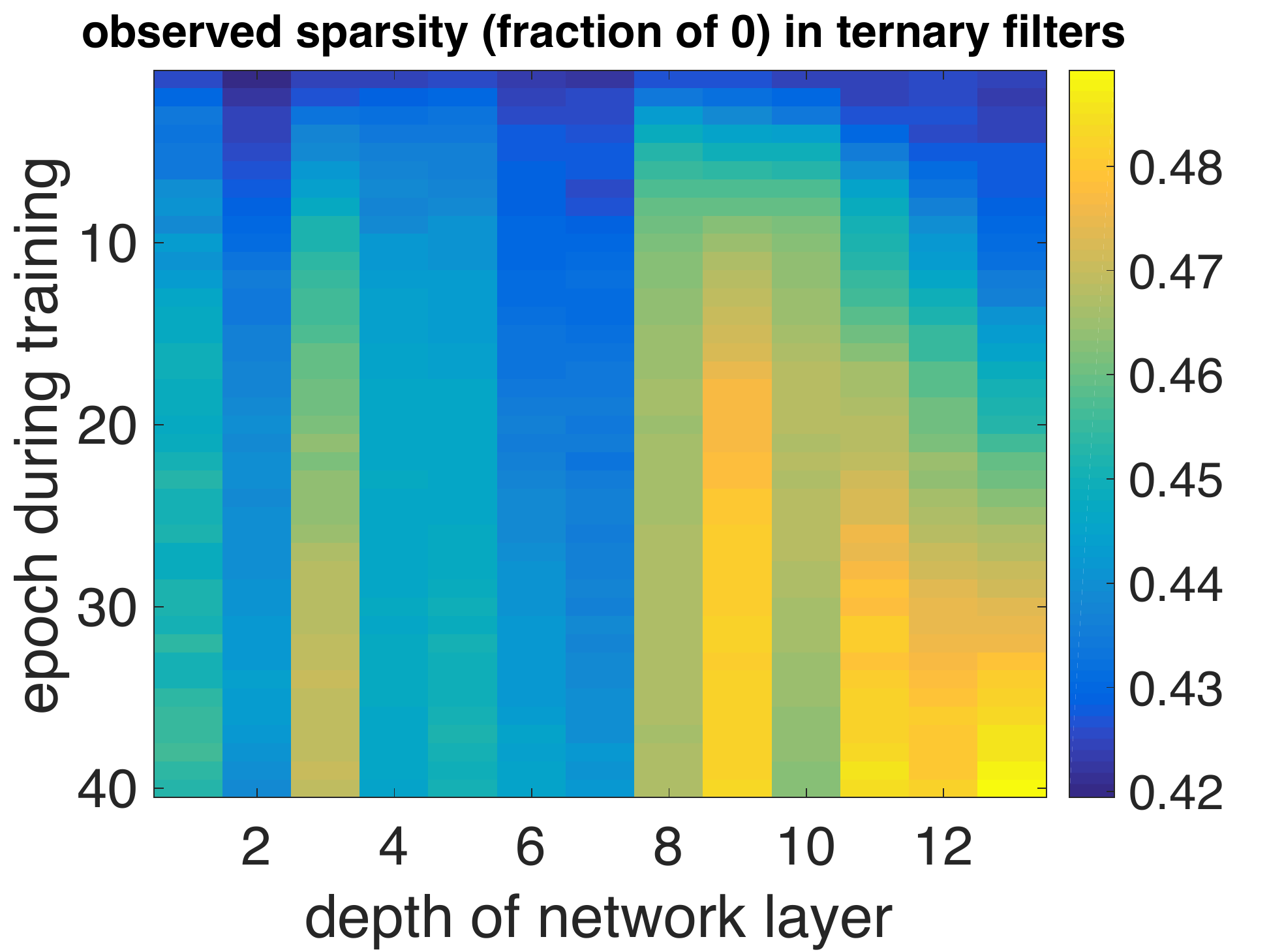}
\end{center}
   \caption{Left: By employing the continuation technique with increasing $\beta$ values during training epochs we can significantly improve the outcome of our trained networks. The ternarised quantisation does thereby no longer affect segmentation quality measured in Dice overlap. Right: The observed sparsity (fraction of zero values) in the trained ternary weights increases throughout the training process. This effect is more pronounced for deeper layers with high parameter counts.}
\label{figDiceLearning}
\end{figure}

When analysing the sparsity of filter weights learned by our model across epochs, shown in Fig.~\ref{figDiceLearning} one can see a tendency to an increase in zero values in later layers and later epochs. In comparison to the number of trainable weights in Table~\ref{tabUnet} it is notable that layers with increased sparsity at the end of training also contain most free parameters. This indicates that the model automatically avoids overfitting and sparsity acts as a regulariser.  The importance of adapting the slope in our ternary hyperbolic tangent nonlinearity during training is clearly shown in Fig.~\ref{figDiceLearning}, where the average Dice is plotted across training epochs. Note, that the evaluation on validation cases always employs ternary convolutions and accordingly quantises activations using Eq.~\ref{eqTernaryAct}.

\section{Conclusion}
We have presented a pioneering approach for ternary convolutions in deep neural networks that relies on both ternarised activations and filter weights. Our work goes beyond previous efforts of binarisation that has often led to severe model degradation. In our experiments, we demonstrated that the \textbf{TernaryNet} maintains the high segmentation quality of the corresponding full precision U-Net (around 71\% Dice for pancreas CT with further potential for improvements), while realising 10$\times$ speed improvements and 15$\times$ lower memory requirements. This is in particular important when executing model inference for image-guided interventions on clinical or mobile computing hardware. A detailed guide to implementation and best practices along with the generality of our approach will help transfer the concept of ternary convolutions to other deep learning applications. We have seen a clear importance of designing a ternary activation that is analytically differentiable based on the underlying hyperbolic tangent nonlinearity as well as using a continuous adaption of its slope during training. This eases the complex training process and results in a high sparsity that is desirable for generalisation and supported by theoretical analysis in literature. When proven in other related fields of computer vision, we strongly believe that quantised networks will have an increasing impact and potentially lead to a wider adaptation of its underlying computational blocks (population counts) in mobile processors.

\subsubsection*{Acknowledgments}
This work was in part supported by the German Research Foundation (DFG) under grant number 320997906

We gratefully thank Nassim Bouteldja who assisted this work by pre-processing the CT scans and helpful discussions. We also acknowledge the great support of NVIDIA with the donation of a Titan Xp GPU used for this research.

\bibliographystyle{plain}

\bibliography{bib2018_arxiv.bib}   

\end{document}